\documentclass[letterpaper, 10 pt, conference]{ieeeconf}  
\IEEEoverridecommandlockouts                              
\usepackage{amsmath}
\usepackage{amssymb}
\usepackage{algorithm} 
\usepackage{algpseudocode}
\usepackage{graphicx}
\usepackage{float}
\usepackage{setspace}
\usepackage{subfigure}
\usepackage{xcolor}
\DeclareMathOperator*{\argmax}{argmax}
\DeclareMathOperator{\sign}{sign}

\renewcommand{\baselinestretch}{0.965}

\newtheorem{problem}{Problem}

\title{\LARGE \bf
Inferring Temporal Logic Properties from Data \\ using Boosted Decision Trees
}

\author{Erfan Aasi$^{1}$, Cristian Ioan Vasile$^{2}$, Mahroo Bahreinian$^{1}$, and Calin Belta$^{1}$
\thanks{*This work was partially supported by the NSF under grants IIS-2024606 and IIS-1723995 at Boston University.}
\thanks{$^{1}$Erfan Aasi, Calin Belta, and Mahroo Bahreinian are with
        Boston University, 
        Boston, MA 02215, USA
        {\tt\small eaasi@bu.edu}, {\tt\small cbelta@bu.edu}, {\tt\small mahroobh@bu.edu}}
\thanks{$^{2}$Cristian Ioan Vasile is with Lehigh University,
        Bethlehem, PA 18015, USA
        {\tt\small cvasile@lehigh.edu}}
}

\begin{document}

\maketitle
\thispagestyle{empty}
\pagestyle{empty}




\begin{abstract}
Many autonomous systems, such as robots and self-driving cars, involve real-time decision making in complex environments, and require  prediction of future outcomes from limited data.
Moreover, their decisions are increasingly required to be interpretable
to humans for safe and trustworthy co-existence.
This paper is a first step towards interpretable learning-based robot control.
We introduce a novel learning problem, called
\emph{incremental formula and predictor learning}, to generate binary classifiers with temporal logic structure from
time-series data.
The classifiers are represented as pairs of Signal Temporal Logic (STL) formulae
and predictors for their satisfaction.
The \emph{incremental} property provides prediction of labels
for prefix signals that are revealed over time.
We propose a boosted decision-tree algorithm
that leverages weak, but computationally inexpensive, learners
to increase prediction and runtime performance.
The effectiveness and classification accuracy of our algorithms are evaluated
on autonomous-driving and naval surveillance case studies.
\end{abstract}

\section{INTRODUCTION}\label{section: introduction}


To cope with the complexity of robotic tasks,
machine learning techniques have been employed to capture their
temporal and logical structure from data.
The drive to provide interpretable learning-based robot control,
where robots can explain their actions and decisions to interact
with humans,
has led to integration of formal methods and temporal logics~\cite{clarke1986automatic}
in machine learning frameworks \cite{asarin2011parametric,jin2015mining,hoxha2018mining,kong2016temporal,bombara2016decision,jha2019telex,vazquez2017logical,neider2018learning,xu2019information,ketenci2019synthesis}.
In this paper, we make the first step towards interpretable, real-time, learning based control, by learning temporal logic properties 
from time-series data and predicting their satisfaction in real-time.


A unique challenge in applying machine learning methods to robotics
is that robots must make decisions with \emph{partially-revealed} information.
Specifically, robots need to take actions before events are completed.
Consider, for example, a self-driving car that must be able to predict, from the behavior of surrounding cars,  whether
it needs to stop for crossing pedestrians 
at the end of a road, when visibility to the end of the road is obstructed. The car should not wait until it is certain of
the presence of pedestrians to take action, as it might not be able to avoid them.  
Moreover, it is also required to explain the decision of whether to stop or not (see Sec.~\ref{sec:motivating_example} for a detailed account of this scenario). 
In this paper, we use \emph{Signal Temporal Logic} (STL) \cite{maler2004monitoring} to provide an interpretable output of the learning process. In our framework, the satisfaction of STL formulas can be predicted from data on events that have yet to finish.



Early methods for mining temporal logic properties from data
require template formulae~\cite{asarin2011parametric,jin2015mining,hoxha2018mining}.
Statistical methods based on simulations of system models are
used to determine the parameters of the formulas.
In~\cite{kong2016temporal}, a general supervised learning framework that could infer both the structure and the parameters of a formula from data is presented. The approach is based on lattice search and parameter synthesis,
which makes it general, but inefficient.
An efficient decision tree-based framework is introduced
in~\cite{bombara2016decision},
and extended to \emph{online learning} in~\cite{bombara2018online},
where constraints on formulae structure are imposed in
exchange for performance.
Other works consider learning temporal logic formulae
from positive examples only~\cite{jha2019telex},
via clustering~\cite{vazquez2017logical} (i.e., unsupervised setting),
using automata-based methods for untimed specifications~\cite{neider2018learning,xu2019information},
and for past-time semantics~\cite{ketenci2019synthesis}.
Once formulae are generated, monitoring methods~\cite{maler2004monitoring}
are available to assess their satisfaction at runtime.
The output of a monitor depends only on the current signal history,
and, most of the time, is inconclusive (i.e., it cannot decide either satisfaction or violation).
None of these methods leverage available data to predict signal labels ahead of time,
and optimize the learned temporal formulae for label prediction.
Moreover, the setting that we consider in this paper differs from \emph{online learning},
where new data instances (signals) are made available one at a time.
We are interested in the case where the data samples themselves are
incomplete, and revealed over time.


In this paper, we introduce the \emph{incremental formula and predictor learning} problem
for binary classification,
where the goal is to generate a pair $(\phi,Pred)$, where $\phi$ is a STL formula and $Pred$ is a prediction function for $\phi$, which together form a classifier of prefix signals.
The classifiers have the required logical structure for interpretability,
while the satisfaction predictors enable evaluation of signals in real time.
To construct such a classifier-predictor pair, we propose a 
boosted decision-tree (BDT) based approach.
Multiple models with weak predictive power and low computational cost are
combined to improve prediction accuracy~\cite{schapire2013boosting}.
The weak learning models are bounded-depth decision trees (DT).
During learning, data samples (signals) are weighted against the DTs
to determine their overall influence on the output of the ensemble,
similar to AdaBoost~\cite{freund1997decision}.
We propose a heuristic inspired by~\cite{domingos2000mining,jin2003efficient},
to decide when to update DT learners as we consider prefix signals
of increasing duration.
Our boosted algorithm can also be applied to the traditional
offline learning.
We evaluate the performance of our algorithms in autonomous-driving
and naval surveillance scenarios.
We show performance gains and increased prediction power compared
to baselines from~\cite{bombara2016decision}.

The main contributions of the paper are:
(a) a novel incremental learning problem that provides predictions of
labels from prefix signals revealed over time;
(b) a new temporal logic inference framework based on BDTs that can be applied for both offline and incremental problems;
(c) an exact MILP encoding for computing STL formulas with minimum impurity at the nodes of a decision tree;
(d) case studies that highlight the prediction and runtime performance
of the proposed learning algorithms in naval docking and autonomous-driving scenarios.

\section{PRELIMINARIES}
\label{section:preliminaries}


Let $\mathbb{R}$, $\mathbb{Z}$, $\mathbb{Z}_{\geq 0}$ denote the sets of real, integer, and non-negative integer numbers, respectively.
With a slight abuse of notation, given 
$a,b\in \mathbb Z_{\geq0}$
we use $[a,b]=\{t\in\mathbb Z_{\geq0}\ |\ a\leq t\leq b\}$.
The cardinality of a set is denoted by $|\cdot|$.
Let $T \in \mathbb{Z}_{\geq 0}$ be a time horizon.
A (discrete-time signal) $s$ is a function $s : [0, T] \to \mathbb{R}^n$
that maps each (discrete) time point $t \in [0, T]$
to an $n$-dimensional vector $s(t)$ of real values.
Each component of $s$ is denoted as $s_j, j \in [1, n]$.
The prefix of $s$ up to time point $t$ is denoted by $s[0{:}t]$.

Signal Temporal Logic (STL) was introduced in~\cite{maler2004monitoring}. 
Informally, STL formulae used in this paper are made of predicates defined over components of real-valued signals in the form 
$s_j(t) \sim \pi$, where $\pi \in \mathbb{R}$ is a threshold, and $\sim \in \{\geq, < \}$, connected using Boolean operators, such as $\lnot$, $\land$, $\lor$, and temporal operators, such as $\mathbf{G}_{[a,b]}$ (always) and $\mathbf{F}_{[a,b]}$ (eventually).
The semantics are defined over signals. For example, formula $\mathbf{G}_{[3,6]}s_3 <1$ means that, for all times 3,4,5,6,
component $s_3$ of a signal $s$ is less than 1,
while formula $\mathbf{F}_{[3,10]}\mathbf{G}_{[0,2]}s_2 > 4$ expresses that at some time between 3 and 10, $s_2(t)$ becomes larger than 4 and remains larger than 4 for at least 3 consecutive time points.

STL has both qualitative (Boolean) and quantitative semantics. We use $s(t)\models\phi$ to denote Boolean satisfaction at time $t$, and $s\models\phi$ as short for $s(0) \models \phi$.
The quantitative semantics is given by a robustness degree (function) \cite{donze2010robust} $\rho(s, \phi, t)$, which captures the degree of satisfaction 
of a formula $\phi$ by a signal $s$ at time $t$. Positive robustness ($\rho(s, \phi, t) \geq 0$) implies Boolean satisfaction $s(t)\models\phi$, while negative robustness ($\rho(s, \phi, t) < 0$) implies violation. For brevity, we use $\rho(s, \phi) = \rho(s, \phi, 0)$.


The {\em horizon} $hrz(\phi)$ of a STL formula $\phi$ is the minimum amount of time necessary to decide its satisfaction. For example, the horizons of the two formulas given above are 6 and 12, respectively.
Weighted STL (wSTL) \cite{mehdipour2020specifying} is an extension of STL that has the same qualitative semantics as STL, but has weights associated with the Boolean and temporal operators, which modulate its robustness degree. In this paper, we restrict our attention to a fragment of wSTL with weights on conjunctions only. For example, the wSTL formula $\phi_1 \land^{\alpha} \phi_2$, $\alpha=(\alpha_1, \alpha_2) \in \mathbb{R}^2_{>0}$,
denotes that $\phi_1$ and $\phi_2$ must hold with priorities $\alpha_1$ and $\alpha_2$. 

Parametric STL (PSTL) \cite{asarin2011parametric} is an extension of STL, where the endpoints $a,b$ of the time intervals in the temporal operators and the thresholds $\pi$ in the predicates are parameters. The set of all possible valuations of all parameters in a PSTL formula $\psi$ is called the parameter space, and is denoted by 
$\Theta$. A particular valuation is denoted by $\theta\in\Theta$, and the corresponding formula by $\psi_\theta$.


\section{PROBLEM FORMULATION} \label{section:problemformulation}

\subsection{Motivating Example} \label{sec:motivating_example}
\begin{figure}[htb]
\centering
\includegraphics[width=0.75\columnwidth]{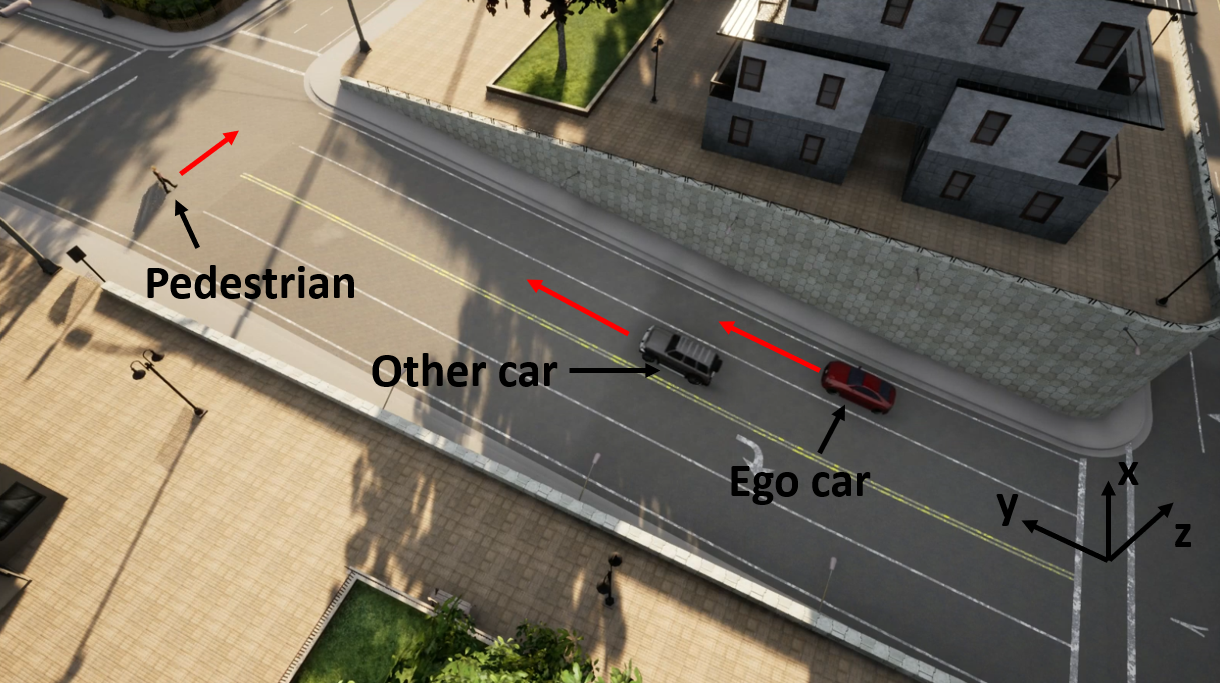}
\caption{Autonomous-driving scenario implemented in CARLA \cite{dosovitskiy2017carla}}
\label{fig:carla_case_study1}
\end{figure}

Consider an autonomous vehicle (referred to as {\em ego}) driving in the urban environment shown in  Fig.~\ref{fig:carla_case_study1}. The scenario contains a pedestrian and another car, which is assumed to be driven by a ``reasonable" human, who obeys traffic laws and common-sense driving rules. Ego and the other car are in different, adjacent lanes, moving in the same direction on an uphill road. We assume that the other car is ahead of ego. The vehicles are headed towards an intersection without any traffic lights. There is an unmarked cross-walk at the end of the road before the intersection. When the pedestrian crosses the street, the other car brakes to stop before the intersection. If the pedestrian does not cross, the other car keeps moving without decreasing its velocity. 

We assume that ego does not have a clear line-of-sight to the pedestrian crossing the street at the intersection, because of the other car and because of the uphill shape of the road. The goal is to develop a method allowing ego 
to infer whether a pedestrian is crossing the street by observing the behavior (e.g., position and velocity over time) of the other car. 

We assume that labeled behaviors (labels indicate whether a pedestrian is crossing or not) are available, and we formulate the problem as a two-class classification problem. First, we 
develop a method that infers an STL formula that classifies whole trajectories (i.e, from their initial time to their final common horizon). Second, motivated by the fact that we want to eventually use such classifiers for real-time control, we propose a method that learns both a formula and a satisfaction predictor during the execution. Using this, we will be able to predict the label of an evolving behavior. For example, ego will be able to make online predictions of whether a pedestrian is crossing by observing the behavior of the other car.  

\subsection{Problem Statement}

Let $C = \{C_n, C_p \}$ be the set of possible classes, where $C_n$ and $C_p$ are the labels for the negative and positive classes, respectively.
We consider a labeled data set with N data samples as $S = \{(s^i, l^i)\}_{i = 1}^{N}$,
where $s^i$ represents the $i^{th}$ signal
and $l^i \in C$ is its label.

\vspace{2mm}
\begin{problem}[{Formula Learning}]\label{probform:non_inc}
Given a labeled data set $S= \{(s^i, l^i)\}_{i = 1}^{N}$,
find an STL formula $\phi$ that minimizes the Misclassification Rate $MCR(\phi)$ defined below:
\begin{equation} \label{eqn:non-incproblem}
    \frac{|\{s^i \mid (s^i \models \phi \, \wedge \, l^i = C_n ) \, \vee \, (s^i \nvDash \phi \, \wedge \, l^i = C_p) \}|}{N}
\end{equation}
\end{problem}
\vspace{2mm}

Problem~\ref{probform:non_inc}, considered in \cite{bombara2016decision,kong2016temporal},
evaluates the classification formula $\phi$ over whole signals from their start
to the horizon $T$.
In contrast, we want to capture prediction performance on prefix signals that have yet
to reach the horizon $T$.
This \emph{incremental} requirement is formalized below.

\vspace{2mm}
\begin{problem}[{Formula and Predictor Learning}] \label{probform:inc}
Given a labeled data set $S = \{(s^i, l^i)\}_{i = 1}^{N}$,
find an STL formula $\phi$ and a satisfaction predictor $Pred(s[0{:}t], \phi)$,
such that the Incremental Misclassification Rate $IMCR(\phi,Pred)$
defined as
\begin{align} \label{eqn:incproblem}
     \frac{1}{T}\sum_{t=1}^{T} &{} \frac{|\{s^i \mid (Pred(s^i[0{:}t], \phi) = C_p \, \wedge \, l^i = C_n ) \, \vee \, }{N} \\ \nonumber
     &{} \frac{(Pred(s^i[0{:}t],\phi) = C_n \, \wedge \, l^i = C_p) \}|}{N}
\end{align}
is minimized, where
$Pred$ maps prefix signals $s^i[0{:}t]$ to a label $C_n$ or $C_p$
that represents the satisfaction prediction of $s^i[0{:}t]$ with respect to $\phi$.
\end{problem}
\vspace{2mm}

\textbf{Discussion:}
In both Problems \ref{probform:non_inc} and \ref{probform:inc}, formula $\phi$ is learnt offline from a batch of labelled signals. The main difference between the learning algorithms solving Problems \ref{probform:non_inc} and \ref{probform:inc} is their usage after learning, i.e, during deployment. A formula $\phi$ learnt using Problem \ref{probform:non_inc} can classify a signal  $s$ only if the signal is known all the way to the final time $T$, i.e. $s[0{:}T]$. However, $(\phi,Pred)$ that solves Problem \ref{probform:inc} classifies signals that evolve in time. In other words, at every $t$, a class prediction is made for $s[0{:}t]$.

To illustrate these ideas, consider the autonomous driving example from Sec. \ref{sec:motivating_example}. 
Data collection for both Problems \ref{probform:non_inc} and \ref{probform:inc} is performed off-line using simulations, real footage from traffic-light cameras, or ego's on-board sensors. During deployment, ego can only use the motion (position, velocity) history
of the other car up to current time $t$ to infer the presence of a pedestrian. Based on the prediction, ego can decide whether to slowdown or stop. Thus, the earlier good predictions are made, the more time the controller has to compute safe plans. Good predictions in this case are captured by the misclassification rate at each time $t$.

\section{SOLUTION TO PROBLEM 1} \label{section:solution1}

In this section, we propose a solution to 
Pb.~\ref{probform:non_inc} based on BDTs.
Our algorithm grows multiple binary DTs based on AdaBoost~\cite{freund1997decision}
that combines weak learners trained on weighted data samples.
Weights represent difficulty of correct classification.
After training a weak learner, the weights of correctly classified samples are decreased and weights of misclassified samples increased.

In Sec.~\ref{sec:boosted_trees} and~\ref{sec:single_trees},
we introduce the BDT construction
and computation of single DT methods, respectively.
We describe methods' meta parameters in Sec.~\ref{sec:meta_params}.
We pose the computation of decision queries associated with trees' nodes
as an optimization problem in Sec.~\ref{subsec: optimization},
and propose a MILP approach to solve them exactly.
Lastly, we explain translation of computed BDTs to STL formulas in Sec.~\ref{sec:translation}.

\subsection{Boosted Decision Tree Algorithm} \label{sec:boosted_trees}

The BDT algorithm in Alg.~\ref{alg:boostedtree} is based on the AdaBoost method.
The algorithm takes as input the labeled data set $S$,
the weak learning model $\mathcal{A}$,
and the number of learners (trees) $B$.
In our approach, the weak learning model $\mathcal{A}$
is the binary DT algorithm Alg.~\ref{alg:dec_tree}.

\begin{algorithm}[htb]
\caption{Boosted Decision Tree (BDT) Algorithm}
\begin{algorithmic}[1]
    \State \textbf{Input:} $S = \{(s^i, l^i)\}_{i = 1}^{N}, \mathcal{A}, B$
    \State \textbf{Output:} final classifier $f_T(\cdot)$
    \State \textbf{Initialize:} $\forall \, (s,l) \in S: \, D_1((s, l)) = 1/|S|$ \label{lst:line:initialize}
    \State for b = 1, $\ldots$, B: \label{lst:line:loop}
    \State \hskip1.5em $\mathcal{A}(S, D_b) \Rightarrow \text{DT classifier} \, f_b$  \label{lst:line:sinlgtree}
    \State \hskip1.5em $\epsilon_b \gets \sum_{(s, l) \in S} D_b((s, l)) \, \cdot \, \mathbf{1}[l \neq f_b(s)]$ \label{lst:line:misclasserror}
    \State \hskip1.5em $\alpha_b \gets \frac{1}{2} \ln{(\frac{1}{\epsilon_b} - 1)} \in \mathbb{R}$ \label{lst:line:treeweight}
    \State \hskip1.5em $D_{b+1}((s,l)) \propto D_b((s,l)) \exp{(-\alpha_b \cdot l \cdot f_b(s))}$  \label{lst:line:weightupdate}
    \State \textbf{return} $f_{total}(s) = \sign(\sum_{b=1}^{B} \alpha_b \cdot f_b(s))$  \label{lst:line:output}
\end{algorithmic}
\label{alg:boostedtree}
\end{algorithm} 

Initially, all data samples are weighted equally (line~\ref{lst:line:initialize}).
The algorithm iterates over the number of trees (line~\ref{lst:line:loop}).
At each iteration, the weak learning algorithm $\mathcal{A}$
constructs a single decision tree $f_b$
based on data set $S$ and current samples' weights $D_b$
(line~\ref{lst:line:sinlgtree}).
Next, the weighted misclassification error of the constructed tree $\epsilon_b$ is computed (line~\ref{lst:line:misclasserror})
that determines the weight of the current tree $\alpha_b$ (line~\ref{lst:line:treeweight}).
At the end of each iteration, the samples' weights are updated and normalized (denoted by $\propto$)
based on the performance of the current tree (line~\ref{lst:line:weightupdate}).
The final output is computed as $f_{total}(\cdot)$ (line~\ref{lst:line:output})
that assigns a label to each data sample
based on the weighted majority vote over all the DTs.
For simplicity, we abuse notation and consider $C_p = 1$
and $C_n  = -1$, such that $f_b(\cdot) \in \{-1, 1\}$ for all $b \in [1, B]$.

\subsection{Single Decision Tree Construction}
\label{sec:single_trees}

Decision trees (DTs)~\cite{breiman1984classification,ripley2007pattern}
are sequential decision models with hierarchical structures.
In our framework, DTs operate on signals with the goal of
predicting their labels.
Their construction is summarized in Alg.~\ref{alg:dec_tree},
and is similar to~\cite{bombara2016decision}.
However, we modify Alg.~\ref{alg:dec_tree} to use
weighted impurity measures (see Sec.~\ref{sec:meta_params})
based on sample weights $D$ that modulate the computation of nodes
(lines~\ref{alg:line:leaf},\ref{alg:line:optimization}),
and partitioning of node data (line~\ref{alg:line:partition}).

\begin{algorithm}[htb]
\caption{Parameterized single decision tree $\mathcal{A}$}
\begin{algorithmic}[1]
    \State \textbf{Meta-Parameters:} $\mathcal{P}, \mathcal{J}, stop$ 
    \State \textbf{Input:} $S = \{(s^i, l^i)\}_{i = 1}^{N}$, $\phi^{path}$, 
    $h$, $D$
    \State \textbf{Output:} sub-tree $\mathcal{T}$ 
    \State \textbf{if} $stop(\phi^{path}, h, S)$ \textbf{then} \label{alg:line:stop}
    \State \hskip1.5em $\mathcal{T} \leftarrow leaf(arg\max_{c \in C} \{p(S, c, D; \phi^{path})\}$) \label{alg:line:leaf}
    \State \hskip1.5em \textbf{return} $\mathcal{T}$
    \State $\phi^*= \underset{\psi \in \mathcal{P}, \theta \in \Theta}{\argmax} \mathcal{J}(S, partition(S, D, \psi_\theta \land \phi^{path}))$ \label{alg:line:optimization}
    \State $\mathcal{T} \leftarrow non\_terminal(\phi^*)$ 
    \State $S_{\top}^*, S_{\bot}^* \leftarrow partition(S, D, \phi^{path} \land \phi^*)$ \label{alg:line:partition}
    \State $\mathcal{T}.left \leftarrow \mathcal{A}(\phi^{path} \land \phi^*, S_{\top}^*, D, h+1)$ \label{alg:line:left}
    \State $\mathcal{T}.right \leftarrow \mathcal{A}(\phi^{path}\wedge \phi^*, S_{\bot}^*, D, h+1)$ \label{alg:line:right}
    \State \textbf{return} $\mathcal{T}$
\end{algorithmic}
\label{alg:dec_tree}
\end{algorithm}

To limit the size and complexity of DTs,
we consider three meta-parameters in Alg.~\ref{alg:dec_tree}:
(1) PSTL primitives $\mathcal{P}$ capturing possible ways to split the data at each node,
(2) impurity measures $\mathcal{J}$ to select the best primitive at each node,
and (3) stop conditions $stop$ to limit the DTs' growth.
The meta-parameters are explained in Sec. \ref{sec:meta_params}.

Alg.~\ref{alg:dec_tree} is recursive,
and takes as input
(1) the set of labeled signals $\mathcal{S}$ reaching the node,
(2) the path formula $\phi^{path}$ from the root to the node (explained in Sec. \ref{sec:translation}),
(3) the depth $h$ from the root to the node, and
(4) the samples' weights $D$ computed in Alg.~\ref{alg:boostedtree}.
First, the stop conditions $stop$ are checked (line~\ref{alg:line:stop}).
If they are satisfied, a single leaf is returned marked with label $c$
(line~\ref{alg:line:leaf})
according to its classification performance (explained in~\ref{subsec:impurity}).
Otherwise, the impurity optimization problem is solved over all primitives
and their valuations to find the best STL primitive $\phi^*$
(line~\ref{alg:line:optimization}), see Sec.~\ref{subsec: optimization}.
Next, data set $S$ is partitioned according to the path formula 
$\phi^{path} \wedge \phi^*$ (line~\ref{alg:line:partition}),
and the algorithm is reiterated for the left and right sub-trees (lines~\ref{alg:line:left}-\ref{alg:line:right}) with their corresponding partitioned data and depth $h+1$.
For simplicity, we group signals with zero robustness with respect to $\phi^*$
in $S^*_\top$, and consider them satisfying.


\subsection{Meta Parameters} \label{sec:meta_params}

\subsubsection{PSTL primitives $\mathcal{P}$} 
The splitting rules at each node are simple PSTL formulas, called {\em primitives},
of two types~\cite{bombara2016decision}: {\em first-order primitives} $\mathcal{P}_1$:  $\mathbf{G}_{[t_0,t_1]} (s_j \sim \pi)$, $\mathbf{F}_{[t_0,t_1]} (s_j \sim \pi)$; and {\em second order primitives} $\mathcal{P}_2$:  $\mathbf{G}_{[t_0,t_1]}(\mathbf{F}_{[0,t_3]} (s_j \sim \pi))$, $\mathbf{F}_{[t_0,t_1]} (\mathbf{G}_{[0,t_3]} (s_j \sim \pi))$.
Decision parameters for $\mathcal{P}_1$ are $(t_0, t_1, \pi)$,
and for $\mathcal{P}_2$ are $(t_0, t_1, t_3, \pi)$.

\subsubsection{Impurity measure $\mathcal{J}$} \label{subsec:impurity}

We use Misclassification Gain (MG) impurity measure \cite{breiman1984classification} as a criterion to select the best primitive at each node. Given a finite set of signals $S$, an STL formula $\phi$, and the subsets of $S$ that are partitioned based on $\phi$ as $S_\top$, $S_\bot=partition (S, \phi)$, we have:
{\small
\begin{align} \label{eqn:mg-impurity}
    &{} MG(S, \{S_\top, S_\bot\}) = MR(S) - \sum_{\otimes \in \{\top, \bot\}} p_\otimes \, MR(S_\otimes)\\ 
    &{} MR(S) = \min (p(S, C_p; \phi) \, , \, p(S, C_n; \phi)), \label{eqn:mg-impurity-mr}
\end{align}}%
where the $p$ parameters are the partition weights of the signals based on their labels and formula $\phi$.
In the canonical definition of impurity measures~\cite{rokach2005},
these are
{\small
\begin{equation*}
    p_\otimes = \frac{|S_\otimes|}{|S|}, \, \otimes \in \{\top, \bot\} , \, \quad p(S, c; \phi) = \frac{|\{(s^i, l^i) \, | \, l^i = c \}|}{|S|}.
\end{equation*}
}%
We extend the robustness-based impurity measures in~\cite{bombara2016decision}
to account for the sample weights $D_b$ from the BDT algorithm.
The boosted impurity measures are defined by partition weights
\begin{equation}
\label{eqn:mg_boosted_weights}
\begin{aligned}
    &{} p_\otimes = \frac{\sum_{(s^i, l^i) \in S_\otimes}^{} D_b((s^i, l^i)) \cdot r(s^i, \phi)}{\sum_{(s^i, l^i) \in S} D_b((s^i, l^i)) \cdot |r(s^i, \phi)|}, \, \, \, \otimes \in \{\top, \bot\} \\
    &{} p(S, c; \phi) = \frac{\sum_{(s^i, l^i) \in S, \, l^i = c}^{} D_b((s^i, l^i)) \cdot |r(s^i, \phi)|}{\sum_{(s^i, l^i) \in S} D_b ((s^i, l^i)) \cdot |r(s^i, \phi)|}
\end{aligned}
\end{equation}
This formulation also works for other types of impurity measures, such as information and Gini gains~\cite{rokach2005}.

\subsubsection{Stop Conditions}
There are multiple stopping conditions that can be considered for terminating Alg.~\ref{alg:dec_tree}. In this paper, we set the stop conditions to either when the trees reach a given depth, or all the signals are correctly classified.

\subsection{Optimization} \label{subsec: optimization}

In the optimization problem for selecting the best primitive in Alg.~\ref{alg:dec_tree},
line~\ref{alg:line:optimization},
we use the MG impurity measure in~\eqref{eqn:mg-impurity} as cost function.
The search space covers all candidate primitives and their valuations.
Due to the nature of the cost function,
nonlinear solvers have been employed in the literature.
Here, we propose a novel Mixed-Integer Linear Program (MILP) formulation
of the problem.
In Sec.~\ref{sec:case_study_naval}, we compare the MILP formulation with
a Particle Swarm Optimization (PSO) \cite{kennedy1995particle} implementation.


\textbf{MILP Formulation:}
We linearize the MG impurity measure using techniques similar  to~\cite{sadraddini2015robust,raman2014model} to formulate a MILP.
Note that the first term in~\eqref{eqn:mg-impurity} only depends on $\phi^{path}$ and is constant with respect to the primitive of the current node.
Thus, maximizing the MG impurity in~\eqref{eqn:mg-impurity} is equivalent to minimizing the second term
{\small
\begin{equation} \label{eqn:simplified-p.MR}
    \sum_{\otimes \in \{\top, \bot\}} p_\otimes \cdot MR(S_\otimes) = p_\top \, MR(S_\top) + p_\bot \, MR(S_\bot),
\end{equation}
}%
where, for simplicity of notation, we write~\eqref{eqn:mg-impurity-mr} as $MR(S_\otimes) = \min (p_{\otimes, p} \, , \, p_{\otimes, n})$
and~\eqref{eqn:mg_boosted_weights} as
\begin{equation*}
   p_\top = \frac{R_\top}{R}, \quad p_\bot = \frac{R_\bot}{R}, \quad
    p_{\otimes, p} = \frac{R_{\otimes, p}}{R_\otimes}, \quad p_{\otimes, n} = \frac{R_{\otimes, n}}{R_\otimes}
\end{equation*}
After simplification, the cost function in~\eqref{eqn:simplified-p.MR} becomes
\begin{equation} \label{eqn:cost-function}
    \mathcal{J} = \min(R_{\top, p}, R_{\top, n}) + \min(R_{\bot, p}, R_{\bot, n}),
\end{equation}
where we eliminated the total robustness $R$ in the denominator
for numerical stability and efficiency.
We write the first element in~\eqref{eqn:cost-function} as
    $
    R_{\top, p} = \sum_{s^i\in S_\top, l^i = C_p} D_b^i \, \, |r^i| = \sum_{s^i \in S, \, l^i = C_p} \max\{D_b^i \, \, r^i, 0\}
    $,
where $r^i = \rho(s^i, \phi_{path} \wedge \phi)$ is the robustness of signal $s^i$,
and $D^i_b$ is its weight.
Similarly, we have $R_{\top, n} = \sum_{s^i \in S, \, l^i = C_n} \max\{D_b^i \, \, r^i, 0\}$, while $R_{\bot, p}$ and $R_{\bot, n}$ are negative sums of min terms.

We encode the robustness values $r^i$ of signals as Mixed-Integer Linear Constraints (MILCs)
\begin{equation*}
    r^i = \min\{\rho(s^i, \phi_{path}), \, \rho(s^i, \phi)\}
    = \min \{q^i, \gamma^i\},
\end{equation*}
where $q^i = \rho(s^i, \phi_{path})$ is a constant with respect to the decision variables $\Theta$ associated with $\phi$, and $\gamma^i = \rho(s^i,\phi)$. 

\textbf{First-order primitives:}
We first assume that the first order primitive is $\phi = \mathbf{G}_{[t_0, t_1]} s^i_\ell \geq \pi$, and then we show how the procedure is applied to the other types of primitives.

We have $\gamma^i = \min_{t \in I} \{s^i_\ell(t) - \pi\} = \min_{t \in I} \{s^i_\ell(t)\} - \pi$, $I=[t_0, t_1]$.
Denote $w^i_{t_0, t_1} = \min_{t\in I} \{s^i_\ell(t)\}$.
Let $\xi_{t_0, t_1} \in \{0, 1\}$ be binary variables associated with integer ranges $[t_0, t_1] \subseteq \mathbb{Z}_{\geq 0}$, $0\leq t_0 \leq t_1 \leq T$, where $T$ is the (common) horizon of the signals.
We encode the robustness $\gamma^i$ as
\begin{align*}
    \gamma^i = \sum_{t_0, t_1} w_{t_0, t_1} \xi_{t_0, t_1} - \pi, \forall i, 
    \text{ and } \sum_{t_0, t_1} \xi_{t_0, t_1} = 1.
\end{align*}

Denote by $P(\{s^i\}_i)$ the procedure to compute
$\theta^* = \argmax_{\theta \in \Theta} \mathcal{J}(\{s^i\}_i)$
for primitive $\phi^{\mathbf{G}, \geq}_\theta = \mathbf{G}_{[t_0, t_1]} s^i_j \geq \pi$.
Since
\begin{align}
    \phi^{\mathbf{G}, <}_\theta &{} = \mathbf{G}_{[t_0, t_1]} s^i_j< \pi = \mathbf{G}_{[t_0, t_1]} (-s^i_j) \geq (-\pi) \\ \nonumber 
    &{} = \mathbf{G}_{[t_0, t_1]} (-s^i_j) \geq \pi' = \phi^{\mathbf{G}, \geq}_{\theta'},
\end{align}
we can solve $\theta^* = \arg\max_{\theta \in \Theta} J(\{-s^i\}_i)$ for primitive $\phi^{\mathbf{G}, <}_\theta$ using the same procedure $P(\cdot)$, i.e., $\theta^* = P(\{-s^i\}_i)$.

Furthermore, we have
\begin{equation}
    \label{eqn:eve-to-alw-transform}
\begin{aligned} 
    \phi^{\mathbf{F}, <}_\theta &= \mathbf{F}_{[t_0, t_1]} s^i_j < \pi = \lnot \mathbf{G}_{[t_0, t_1]} s^i_j \geq \pi = \lnot \phi^{\mathbf{G}, \geq}_{\theta},\\
    \phi^{\mathbf{F}, \geq}_\theta &= \mathbf{F}_{[t_0, t_1]} s^i_j \geq \pi = \lnot \mathbf{G}_{[t_0, t_1]} s^i_j < \pi = \lnot \phi^{\mathbf{G}, <}_{\theta},
\end{aligned}
\end{equation}
which means that, if we have already computed the parameters for the primitives
$\phi^{\mathbf{G}, \geq}_\theta$ and $\phi^{\mathbf{G}, <}_\theta$,
we do not need to solve separate problems to find the parameters for
$\phi^{\mathbf{F}, <}_\theta$ and $\phi^{\mathbf{F}, \geq}_\theta$
as they are related to each other according to~(\ref{eqn:eve-to-alw-transform}).

\textbf{Second-order primitives:} The MILC encoding of second-order primitives is similar to the first-order ones.
Consider the primitive $\phi = \Box_{[t_0, t_1]}(\lozenge_{[0, t_3]} s^i_j \geq \pi)$, and denote $w^i_{t_0, t_1, t_3} = \min_{t\in [t_0, t_1]} \{\max_{t' \in [t, t+t_3]}  \{s^i_j(t')\}\}$.
Let $\xi_{t_0, t_1, t_3} \in \{0, 1\}$ be binary variables associated with integer ranges $[t_0, t_1]([0, t_3]) \subseteq \mathbb{Z}_{\geq 0}$, $1 \leq t_3 \leq T$, $0\leq t_0 \leq t_1 \leq T - t_3$, where $T$ is the (common) horizon of the signals.
We encode the robustness $\gamma^i$ as
\begin{align*}
    \gamma^i = \sum_{t_0, t_1, t_3} w_{t_0, t_1, t_3} \xi_{t_0, t_1, t_3} - \pi, \forall i,
    \text{ and }
    \sum_{t_0, t_1, t_3} \xi_{t_0, t_1, t_3} = 1
\end{align*}
The encoding of the other second-order primitives follows similarly to the procedure for first-order primitives.

\subsection{Decision trees to formulas}
\label{sec:translation}
We use the method from~\cite{bombara2016decision},
shown in Alg.~\ref{alg:tree2stl}, to convert a DT to an STL formula.
The algorithm is invoked from the root, and builds the formula that
captures all branches that end in leaves marked $C_p$.

\begin{algorithm}[h] 
\caption{Tree to STL formula - $Tree2STL(\cdot)$}
\begin{algorithmic}[1]
    \State \textbf{Input:} node $\mathcal{T}$ \qquad
    \textbf{Output:} STL formula $\phi$
    \State \textbf{if} $\mathcal{T}$ is a leaf and is marked with $C_p/C_n$ \textbf{then} 
    \State \hskip1.5em \textbf{return} $\top / \bot$
    \State $\phi_l = (\mathcal{T}.prim \wedge Tree2STL(\mathcal{T}.left))$
    \State $\phi_r = (\neg \mathcal{T}.prim \wedge Tree2STL(\mathcal{T}.right))$
    \State \textbf{return} $\phi = \phi_l \vee \phi_r$ 
\end{algorithmic}
\label{alg:tree2stl}
\end{algorithm} 

The BDT method returns a set of formulas and associated weights $\{(\phi_b, \alpha_b)\}_{b=1}^{B}$ used
in the weighted majority vote classification scheme.
The STL formula $\Phi = \bigwedge_{b} \phi_b$
is the overall output formula.
However, if we use wSTL \cite{mehdipour2020specifying},
we can express $\Phi = {\bigwedge_{b}}^{\alpha_b} \phi_b$,
where the weights are part of the formula.


\begin{figure}[h]
\centering
\includegraphics[width=0.8\columnwidth]{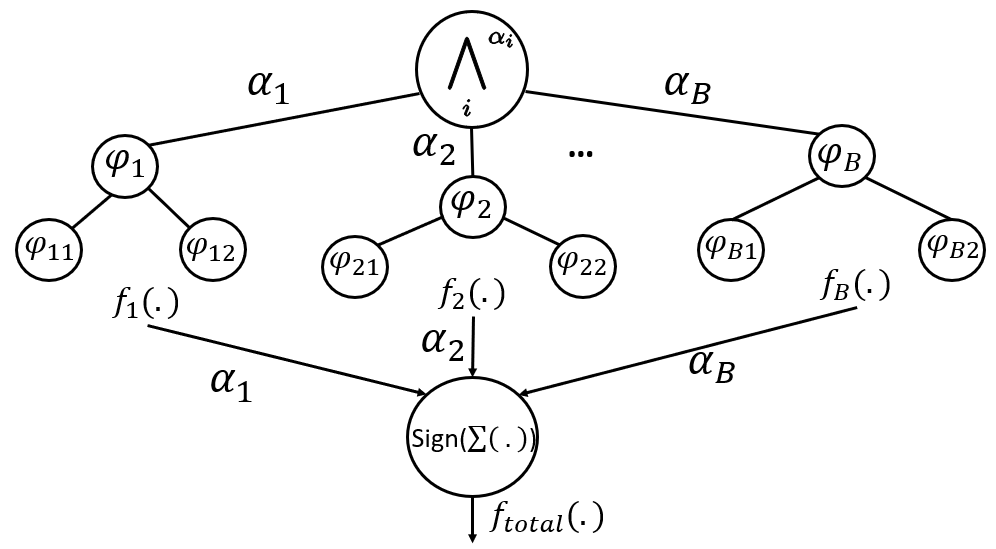}
\caption{Illustration of boosted decision tree algorithm. The final classifier $f_{total}$ is the sign of the weighted sum of the DTs' outputs taken as the label.
The DTs and weights $\alpha_i$ are translated to
a wSTL formula.}
\label{figurelabel}
\end{figure}

\section{SOLUTION TO PROBLEM 2}
\label{section:solution2}


Pb. \ref{probform:inc} poses two main challenges:
(1) ensuring prediction performance on prefix signals completed over time,
and (2) avoiding the explosion in computation time due to the incremental requirement.
We address both problems by adapting the boosted tree method from
Sec.~\ref{section:solution1}.
As a departure from boosted methods that consider weak learners trained
on the same data set,
we create DTs on subsets of prefix signals of increasing duration.
Moreover, we adapt an idea from online learning to decide
when to add new DTs to the set of weak learners.


It is important to note the difference between
the incremental learning problem Pb.~\ref{probform:inc}
and online learning problems~\cite{bombara2018online}.
In the latter, the goal is to perform training when
samples (labeled signals in this paper) are provided one at the time.
Much of the literature is focused on ensuring provable performance 
with respect to the batch case, where all data is available at once.
In contrast, the incremental learning problem Pb.~\ref{probform:inc}
is a supervised binary classification problem, where the samples are all given from the start, but they are evaluated over time.
Thus, when deployed, classifiers are able to predict signal labels
ahead of their completion time.


\subsection{Incremental Boosted Decision Trees} \label{sec:inc_boosted}
In Alg.~\ref{alg:inc_boostedtree}, we propose a heuristic BDT algorithm for Pb.~\ref{probform:inc}. 
The inputs are the entire batch of labeled data $S$ and the single decision tree $\mathcal{A}$ (see Alg.~\ref{alg:dec_tree}).
The output is a set $\mathcal{O}$ of trained trees and their weights.
In contrast to Alg.~\ref{alg:boostedtree}, the number of decision trees $B$ is not a tunable parameter, but is determined during training. 
After initialization of data weights (line~\ref{inc:initial}),
the algorithm iterates over each timepoint up to the signals' horizon $T$ (line~\ref{inc:loop}),
computes $S^{par}_t$ as the set of prefix signals up to time point $t$ and their labels (line~\ref{inc:partial}).
At line~\ref{inc:constraint}, we check using $createTree(\,)$
whether $S^{par}_t$ is informative enough to create a tree
that achieves good prediction performance for the next timepoints,
as the signal traces get complete over time.
We detail $createTree(\,)$ below.
The rest of Alg.~\ref{alg:inc_boostedtree} (lines~\ref{inc:single_tree}-~\ref{inc:update_data_weights})
proceeds as Alg.~\ref{alg:boostedtree}, and returns the trained trees and weights $\mathcal{O}$ (line~\ref{inc:append}).
Note that our algorithm avoids the naive, inefficient approaches
of training based on all prefixes at once,
and generating DTs at every time step.


\begin{algorithm}[h] 
\caption{Incremental Boosted Decision Tree Algorithm}
\begin{algorithmic}[1]
    \State \textbf{Input:} $S = \{(s^i, l^i)\}_{i = 1}^{N} , \, \mathcal{A}$
    \State \textbf{Output:} $\mathcal{O} = \{(f_b(\cdot), \alpha_b)\}_{b = 1}^{B}$
    \State \textbf{Initialize:} $\forall \, (s,l) \in S: \, D(s, l) = 1/|S|$ \label{inc:initial}
    \State for t = 1, ..., T:  \label{inc:loop}
    \State \hskip1.5em $S^{par}_t \gets \{(s^i[0{:}t], \, l^i)\}_{i = 1}^{N}$ \label{inc:partial}
    \State \hskip1.5em if $createTree (S^{par}_t) = True$: \label{inc:constraint}
    \State \hskip1.5em \hskip1.5em $\mathcal{A}(S^{par}_t, D) \Rightarrow \, f$ \label{inc:single_tree}
    \State \hskip1.5em \hskip1.5em $\epsilon \gets \sum_{(s, l) \in S^{par}_t} D((s, l)) \cdot \mathbf{1}[l \neq f(s)]$
    \State \hskip1.5em \hskip1.5em $\alpha \gets \frac{1}{2} \ln{(\frac{1}{\epsilon} - 1)} \in \mathbb{R}$ 
    \State \hskip1.5em \hskip1.5em $D((s,l)) \propto D((s,l)) \exp{(-\alpha \cdot l \cdot f(s))}$  \label{inc:update_data_weights}
    \State \hskip1.5em \hskip1.5em $\mathcal{O}.append((f, \alpha))$ \label{inc:append}
    \State \textbf{return} $\mathcal{O}$  \label{inc:output}
\end{algorithmic}
\label{alg:inc_boostedtree}
\end{algorithm} 

\subsection{Single Incremental Tree creation}
\label{sec:inc_tree_create}

We propose procedure $createTree(\,)$ that
checks whether generating a DT from $S_t^{par}$
leads to increased prediction performance.
Let $\mathcal{P} \in \{\mathcal{P}_1, \mathcal{P}_2\}$ be the set of primitives,
$\phi^*_\psi$ the STL formula with optimal valuation for $\psi \in \mathcal{P}$,
and $\psi^*$ the optimal primitive with respect to $MG$ for $S^{par}_t$.
We say that $\psi \in \mathcal{P} \setminus \{\psi^*\}$ is informative if
    $MG(S^{par}_t, \phi^*_{\psi^*}) - MG(S^{par}_t, \phi^*_\psi) > \epsilon(S^{par}_t, \delta)$,
where $(1 - \delta) \in (0, 1)$ is interpreted as
the probability that $\psi^*$ is better than $\psi$
with respect to the complete dataset $S$,
and $\epsilon(S^{par}_t, \delta) > 0$ is set as in~\cite{rutkowski2014new}.
If all primitives $\psi \in \mathcal{P} \setminus \{\psi^*\}$ are informative,
$createTree(\,)$ returns true.
The test is based on a procedure from online 
learning~\cite{bombara2018online,domingos2000mining},
but applied to prefix datasets $S^{par}_t$ instead.
Since the setup differs, the theoretical results~\cite{rutkowski2014new}
do not necessary hold, but we show the prediction performance
empirically in Sec.~\ref{section:case-studies}.

\subsection{Satisfaction Predictors} 
Here we explain how to build predictors that map prefix signals to labels in $\{C_p, C_n\}$.
Let $\mathcal{O} = \{(f_b(\cdot), \alpha_b)\}_{b = 1}^{B}$ be the set of DTs and
their weights generated by Alg.~\ref{alg:inc_boostedtree}.
We translate all DTs to equivalent STL formulae $\phi_b$
with horizons $hrz(\phi_b)$, using Alg.~\ref{alg:tree2stl}.
The \emph{predictor} associated with $\mathcal{O}$ is
$\mathcal{M} = \{(M_b(\cdot), hrz(\phi_b))\}_{b = 1}^{B}$,
where $M_b(\cdot)$ is a monitor~\cite{maler2004monitoring} for $\phi_b$.
The output of the predictor is computed based on the weighted voting of monitors'
outputs with random choices in inconclusive cases.
Formally, let $t \leq T$ be the current time step, where $T$ is the signals' horizon.
The set of active monitors is $\mathcal{M}^{act}_t = \{M_b\mid hrz(\phi_b) \leq t\}$.
Thus, the predictor's output is $Pred(s[0{:}t], \Phi) =
\sign\Big(\sum_{M_b \in \mathcal{M}^{act}_t} \alpha_b \cdot M_b(s[0{:}t], \phi_b)\Big)$
when $\mathcal{M}^{act}_t \neq \emptyset$,
and $Pred(s[0{:}t], \Phi) \sim Ber(0.5)$,
where $Ber(0.5)$ are i.i.d. random uniform Bernoulli trials,
i.e., fair coin tossing,
and $\Phi = {\bigwedge_{b}}^{\alpha_b} \phi_b$, see Sec.~\ref{sec:translation}.
Note that the proposed predictor is tailored for the output
of Alg.~\ref{alg:inc_boostedtree},
and may not work with Alg.~\ref{alg:boostedtree},
because formulae are not learned to capture prefix signals.
This is shown in the next section.

\section{CASE STUDIES}
\label{section:case-studies}


We demonstrate the effectiveness and computational advantages of our methods with two case studies.
The first one is the autonomous-driving scenario from Sec.~\ref{sec:motivating_example},
where we focus on the interpretability and the prediction power of our algorithms.
The second is a naval surveillance problem from~\cite{kowalska2012maritime,kong2016temporal},
where the emphasis is on prediction and runtime performance, compared to~\cite{bombara2016decision}
as baseline. We also compare PSO and MILP based implementations in the naval case study.


%

\textbf{Evaluation.}
In both cases, we evaluate our offline and incremental algorithms, based on
prediction and runtime performance using testing phase MCR and execution time as metrics,
computed using k-fold cross validation and PSO.
We show the prediction  power of the incremental Alg.~\ref{alg:inc_boostedtree}
in comparison with the offline method in Alg.~\ref{alg:boostedtree} as baselines
that we trained on whole signals (non-incremental).
Both methods are tested incrementally on prefix signals.


In the test phase of the k-fold cross validation,
the MCR and IMCR are computed based
on~\eqref{eqn:non-incproblem}, and~\eqref{eqn:incproblem}, respectively.
However, in the evaluation of the incremental learning method,
the monitors may return inconclusive results (neither satisfaction nor violation).
In such cases, we assign a value of 0.5 to MCR (or IMCR).
We make this convention, because we force offline algorithms to make
choices based on i.i.d. fair coin tosses when inconclusive.
Similarly, when our predictors are inconclusive we assign 0.5,
since the choice is stochastic.

Qualitatively, we show the interpretability of our framework using the
generated formulas and their meaning in the case studies.
Moreover, in the implementation of incremental learning in both case studies,
we interpret how early prediction power is critical for downstream control and decision making.

The execution times are measured based on the system's clock.
All computations are done in Python 3 on an Ubuntu 18.04 system
with an Intel Core i7 @3.7GHz and 16GB RAM.
We use Gurobi~\cite{optimization2014inc} to solve the MILPs.

We report results in tables that show
number of folds (F),
tree depth (D),
number of BDTs (\# of T),
mean testing MCR (MCR M),
standard deviation of testing MCR (MCR S),
and execution time (R).




\subsection{Autonomous Driving Scenario}
\label{sec:carla_case_study}

This case study is implemented in the CARLA simulator~\cite{dosovitskiy2017carla} (see Fig.~\ref{fig:carla_case_study1}).
The pedestrian and the other car are both on the same side of the street.
The acceleration of both cars is constant, and smaller for ego than the other vehicle.
The initial positions and accelerations of the cars are initialized such that there is always a minimum safe distance between them.


The simulation of this scenario ends whenever ego gets closer than 8$m$ to the intersection.
We assume ego is able to estimate the relative position and velocity of the other car.
The cars move uphill in the $y-z$ plane of the coordinate frame,
towards positive $y$ and $z$ directions,
with no lateral movement in the $x$ direction.
We collected 300 signals with 500 uniform time-samples per trace,
where 150 were with and 150 without pedestrians crossing the street.

\begin{figure}[htb]
    \centering
    \subfigure[]
    {\includegraphics[width=0.3\columnwidth]{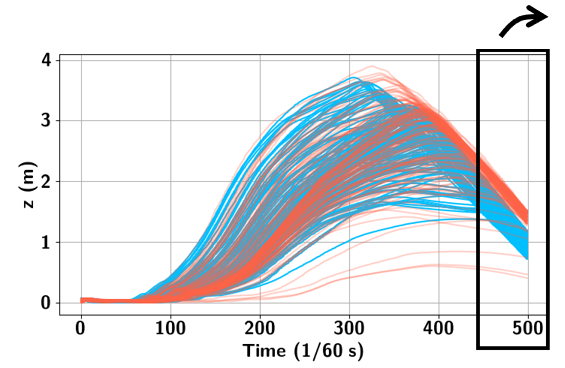} \label{fig:carla_all_z}}
    \subfigure[]
    {\includegraphics[width=0.65\columnwidth]{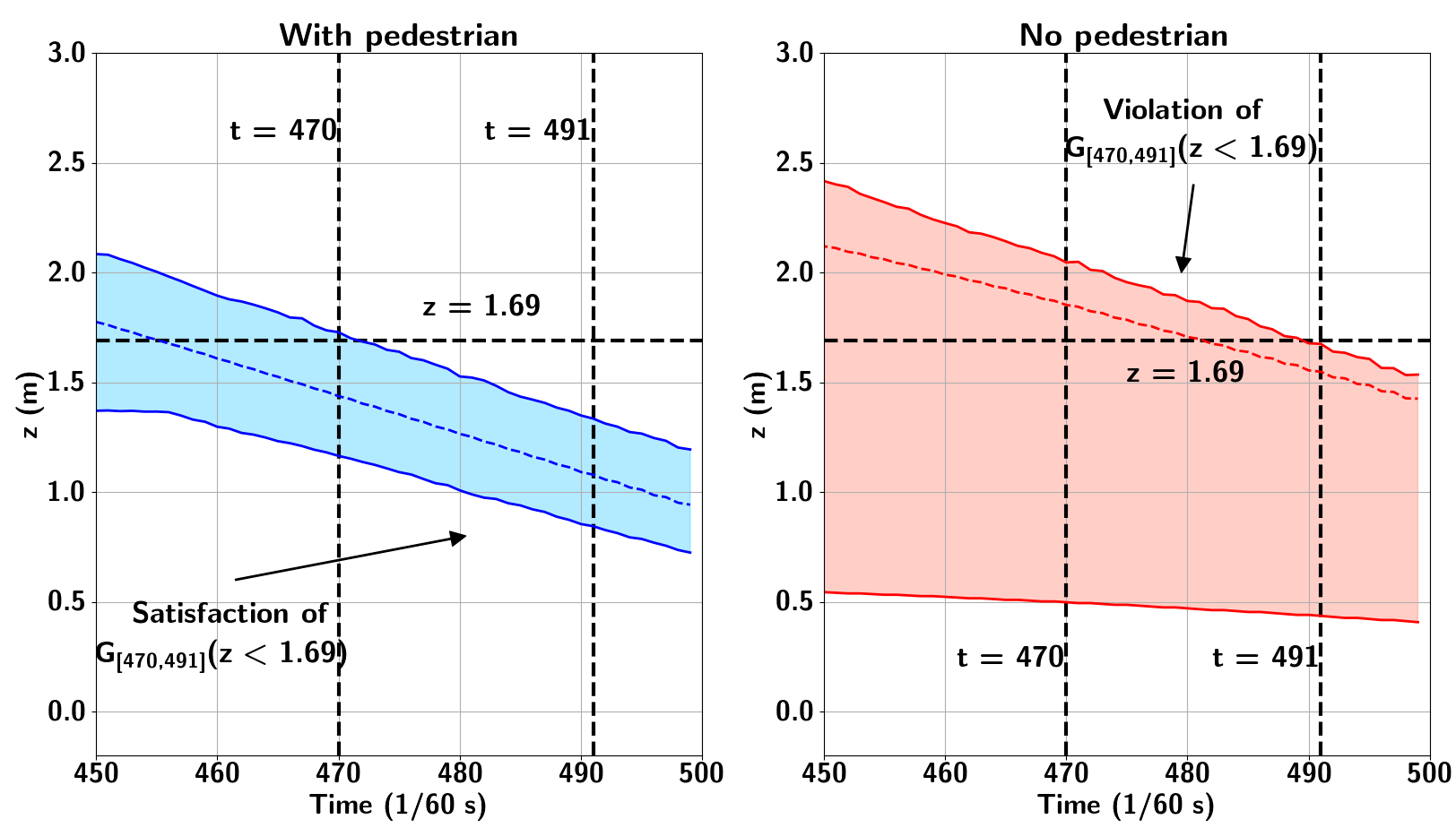} \label{fig:carla_z}}
    \subfigure[]
    {\includegraphics[width=0.3\columnwidth]{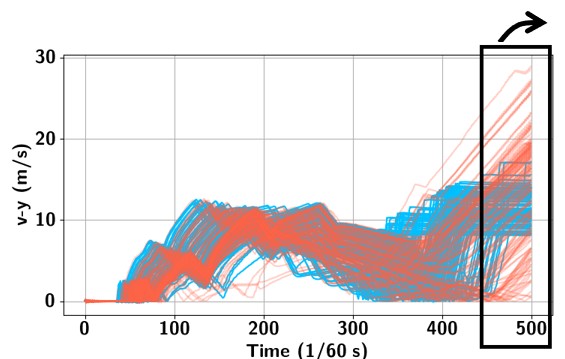} \label{fig:carla_all_vy}}
    \subfigure[]
    {\includegraphics[width=0.65\columnwidth]{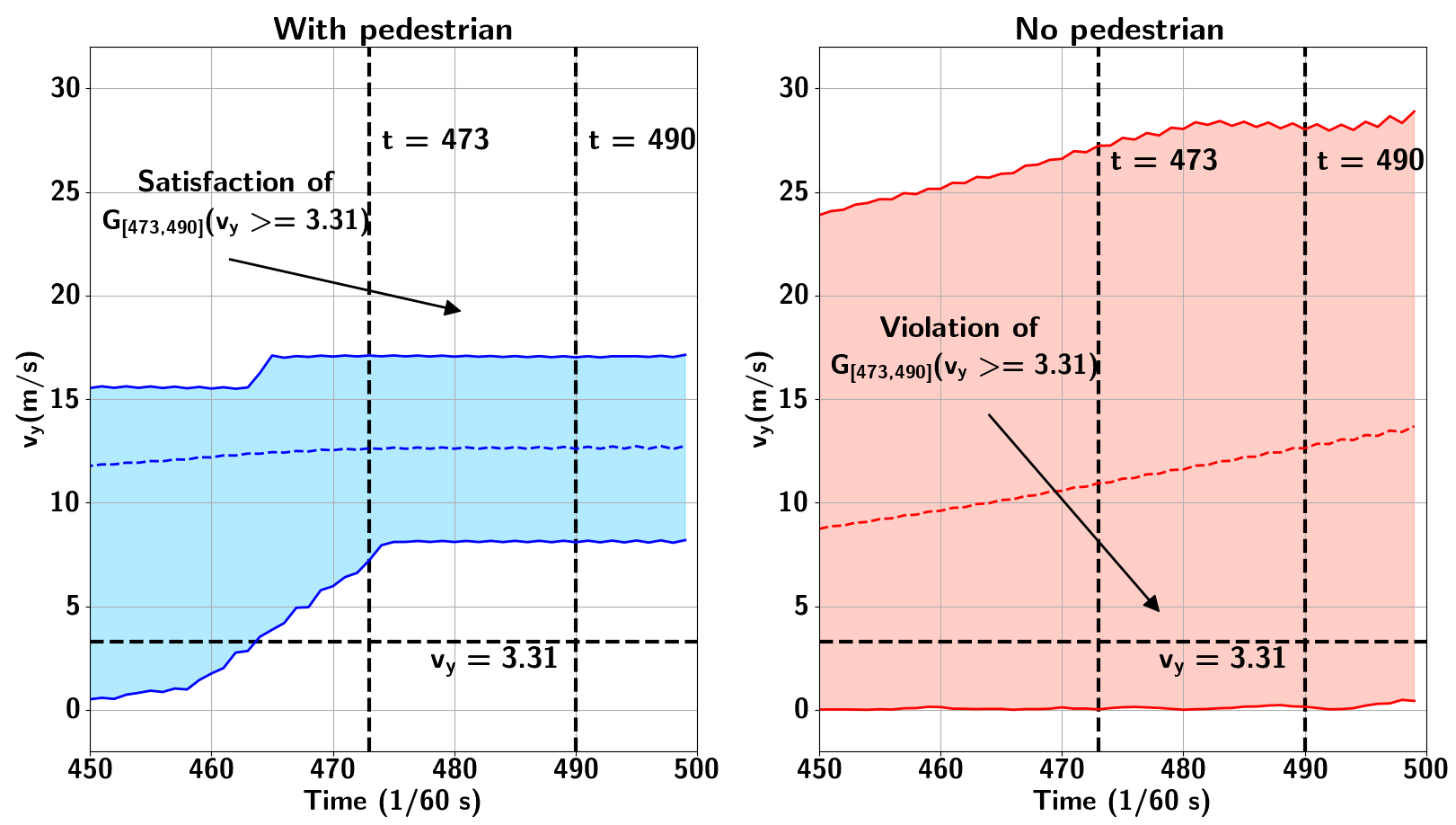} \label{fig:carla_vy}}
    \subfigure[]
    {\includegraphics[width=0.3\columnwidth]{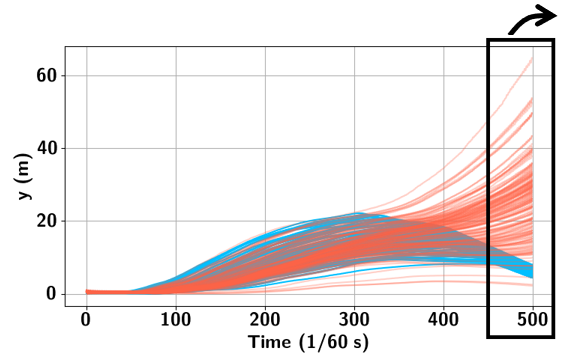} \label{fig:carla_all_y}}
    \subfigure[]
    {\includegraphics[width=0.65\columnwidth]{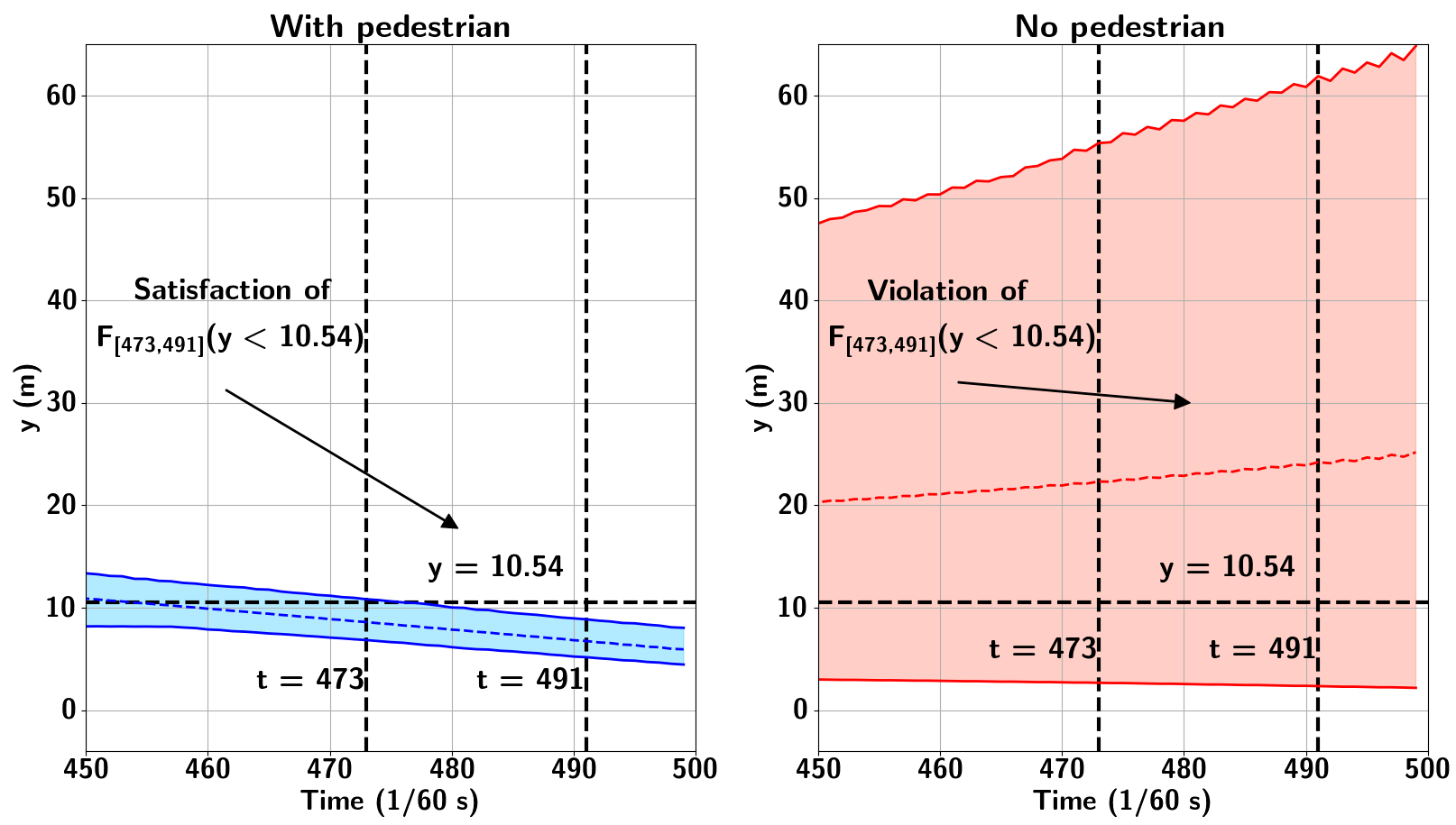} \label{fig:carla_y}}
    \caption{
    Representation of the entire signals (left column) and the last 50 timepoints of the signals (middle and right column) in the autonomous-driving scenario.
    The blue signals on the left and middle column, and the red signals on the left and right column correspond to scenarios with and without pedestrians
    crossing the street, respectively.
    All signals are the absolute values of relative distances and velocities.
    \subref{fig:carla_z}, \subref{fig:carla_vy}, and \subref{fig:carla_y}
    show the distance in $z$-direction, velocity in $y$-direction, and distance in $y$-direction, respectively.
    In the middle and right column, the mean of signals and their range (min, max) at each timepoint are shown
    as colored dash-lines and regions, respectively.
    }
    \label{fig:carla_signals}
\end{figure}

\begin{table}[htb]
\scriptsize
\caption{}
\vspace{-4mm}
\label{table:results}
\begin{center}
\begin{tabular}{|c||c||c||c||c||c||c|}
\hline
Case & F & D & \# of T & MCR M (\%) & MCR S (\%) & R \\
\hline
Car & 3 & 1 & 3 & 1.66 & 0.47 & 6h 42m \\
\hline
Car & 5 & 2 & 1 & 1 & 2 & 4h 7m \\
\hline
Naval & 5 & 3 & 4 & 0.35 & 0.5 & 5h \\ 
\hline 
Naval & 10 & 3 & 3 & 0.65 & 1.05 & 9h 47m \\ 
\hline
\end{tabular}
\end{center}
\vspace{-4mm}
\end{table}

\textbf{Classification and Prediction Performance:}
The evaluation of our offline algorithm Alg.~\ref{alg:boostedtree} is done with 3- and 5-fold cross validation, and the results are reported in the first two rows of Table~\ref{table:results}.
Then, we assess the incremental algorithm Alg.~\ref{alg:inc_boostedtree}, with 3-fold cross validation and depth two for the trees. 

\begin{figure}[htb]
\centering
\includegraphics[width=0.85\columnwidth]{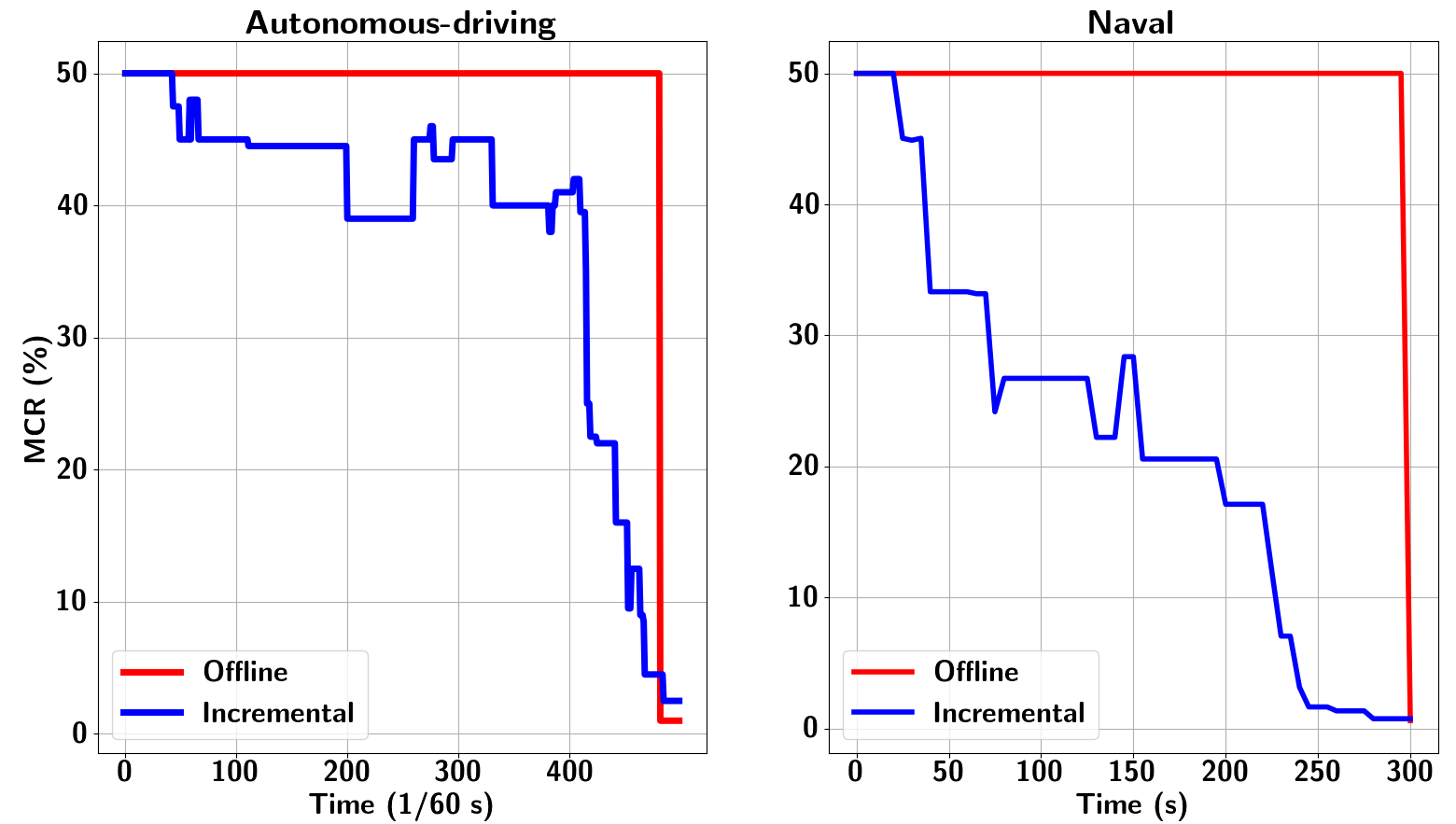}
\caption{
Evaluation of the incremental learning method. The left and right plots show the performance in autonomous-driving and naval cases, respectively.
Both figures show that the incremental learning algorithm (blue)
has better prediction performance than the offline learning algorithm (red) evaluated incrementally in the test phase. The IMCR is proportional to the area under the curves.
}
\label{fig:inc}
\end{figure}

The assessment of incremental learning is shown in Fig.~\ref{fig:inc} on the left.
For the offline method (red curve), since the learnt formula's horizon requires the whole trace of the signals,
it is not able to predict labels for the signals and the MCR is 50$\%$ (IMCR is 49.19$\%$).
In the incremental case (blue curve), after the first $\sim$50 timepoints,
the algorithm is able to predict labels for the signals and the MCR drops below 50$\%$ (IMCR is 22.08$\%$).
Since the representative information within signals is mostly towards end timepoints (see Fig.~\ref{fig:carla_signals}),
the MCR does not change significantly until timepoint 400,
where it drops drastically and the predictive power of the incremental model increases.

\textbf{Interpretability:}
For brevity, we report a learnt formula in one split of the 5-fold cross validation with the offline algorithm: $\phi = (\phi_1 \wedge \phi_2) \vee (\neg \phi_1 \wedge \phi_3)$, where $\phi_1 = \mathbf{G}_{[470, 491]} (z < 1.69)$, $\phi_2 = \mathbf{G}_{[473, 490]} (v_y \geq 3.31)$, and $\phi_3 = \mathbf{F}_{[473, 491]} (y < 10.54)$.
From Fig.~\ref{fig:carla_signals}, we can see how the learnt formula captures the key features of the signals located mainly near the end timepoints.
In English, the formula states that, when a pedestrian crosses the street, the relative distance of the vehicles in the $z$-direction is below 1.69$m$ during nearly the last 30 timepoints, and the relative velocity in the $y$-direction is greater than or equal to 3.31$m/s$.

This is reasonable since the other vehicle is stopped before the intersection, and ego is getting close to the other vehicle.
In the incremental learning evaluation shown in Fig.~\ref{fig:inc},
we can see that the MCR of the incremental algorithm is better than the offline method,
which enables ego to decide whether a pedestrian is crossing the street before reaching the intersection.
Thus, ego has more time to react and behave safely.

\textbf{Runtime Performance:}
The execution time of the offline learning implementations are mentioned in the last column of Table~\ref{table:results}.
The runtime for training of the incremental algorithm is 5 hour and 19 minutes, while for the offline algorithm it is about 7 minutes.
Although the incremental method is slower, it is able to predict labels on the fly with better accuracy compared to the offline learning method.

\subsection{Naval Surveillance}
\label{sec:case_study_naval}

\begin{figure}[h]
\centering
\includegraphics[width=0.55\columnwidth]{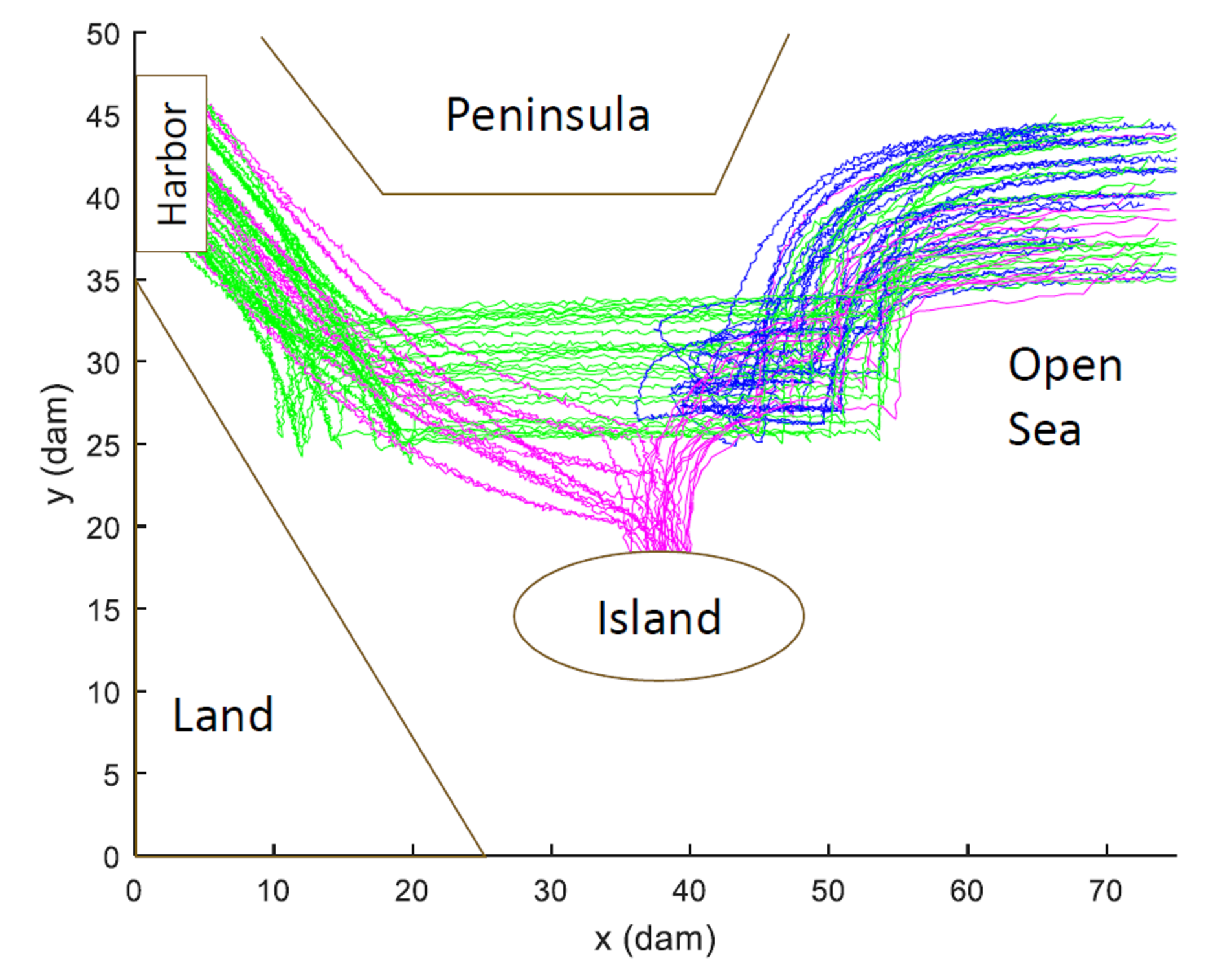}
\caption{Naval surveillance scenario \cite{kong2016temporal}, where normal trajectories are shown in green, and anomalous ones are shown in blue and magenta.}
\label{fig:naval_case_study}
\end{figure}

The naval surveillance problem was proposed in~\cite{kong2016temporal} based on the scenarios
from~\cite{kowalska2012maritime} with the goal of detecting anomalous behavior of vessels
from their trajectories. Normal trajectories belong to vessels approaching from open sea and heading directly to a harbor.
Anomalous trajectories belong to vessels that either veer to another island first and then head to the same harbor,
or they approach other vessels midway to the harbor and then veer back to the open sea (see Fig.~\ref{fig:naval_case_study}).
More details can be found in~\cite{kong2016temporal}.


\textbf{Classification and Prediction Performance:}
We evaluate the offline algorithm Alg.~\ref{alg:boostedtree} with 5- and 10-fold cross validations and the results are represented in the last two rows of Table~\ref{table:results}.
In~\cite{bombara2016decision} for 5-fold cross validation and using only first-order primitives and trees with depth $\leq$ 4, they obtained the mean MCR of 0.4$\%$,
and for 10-fold cross validation using only second-order primitives and trees with depth $\leq$ 3, they reached mean MCR of 0.7$\%$.
Comparing our results in Table~\ref{table:results} with~\cite{bombara2016decision}, 
our algorithm's relative improvement against the MCR of \cite{bombara2016decision} is 12$\%$ in 5-, and 7$\%$ in 10-fold cross validations.

The performance of models trained using the offline Alg.~\ref{alg:boostedtree}
and incremental Alg.~\ref{alg:inc_boostedtree} algorithms is shown in Fig.~\ref{fig:inc} on the right,
where the evaluation was performed incrementally during testing.
Similar to the autonomous-driving case, the offline learned models require full signals to compute labels (IMCR is 48.24$\%$),
while the incremental models can predict early with much better accuracy (IMCR is 38.34$\%$).

\textbf{Interpretability:}
The right plot in Fig.~\ref{fig:inc} shows that the most important information within the signals
is around the middle of the time window.
The prediction power of the incremental model increases and its MCR decreases almost linearly.
This enables the prediction of anomalous behaviors much earlier than the offline learning method,
which provides the opportunity to mitigate the possible damages before it is too late.

\textbf{Runtime Performance:}
In the offline learning framework, in \cite{bombara2016decision}, the runtime was 16 minutes per split (1h20m in total) for the 5-fold, and 4 hour per split (40h in total) for the 10-fold cross validation.
Comparing our results in Table~\ref{table:results} with \cite{bombara2016decision}, and considering the fact that we are using the BDT method, which is a slower algorithm compared to growing a single decision tree, we can conclude that our algorithm scales better with the number of trees.
In the incremental learning evaluation, the execution time of the incremental method is 1h7m and the runtime of the offline method is 8 minutes. This confirms the fact that the incremental learning algorithm is slower, but considering its prediction performance mentioned earlier, it is more accurate than the offline learning method. 

\textbf{Optimization and Overfitting:}
Here we evaluate the proposed MILP formulation in Sec.~\ref{subsec: optimization}, in the offline learning framework using first-order primitives, and compare its results with PSO-based implementations with same initialization for parameters. The results are shown in the table below. We can see that with same set of parameters for each method, the MILP method is slower than the PSO. Moreover, since the MILP finds the globally optimal solution at each node, in trees with higher depth, the MILP implementation is at risk of overfitting. This can be interpreted from the table in the sense that in first and third row with depth = 1, both PSO and MILP have similar mean MCR in training and test phase, but in the second and last row when the trees get more deep (depth = 3), the MILP has mean MCR of 1.45$\%$ and PSO has mean MCR of 1.75$\%$ in the training phase and MILP gets less accurate in the test phase than the PSO.

\begin{table}[h]
\scriptsize
\vspace{-1mm}
\label{table:milp_results}
\begin{center}
\begin{tabular}{|c||c||c||c||c||c|}
\hline
Method & F & D & \# of T & MCR M (\%) & R \\
\hline
MILP & 3 & 1 & 1 & 22.40 & 1h 7m \\
\hline
MILP & 2 & 3 & 1 & 6.90 & 1h 9m \\ 
\hline
PSO & 3 & 1 & 1 & 22.70 & 1m 4s \\
\hline
PSO & 2 & 3 & 1 & 2.2 & 1m 35s \\ 
\hline
\end{tabular}
\end{center}
\vspace{-5mm}
\end{table}


\section{CONCLUSION} \label{section:conclusion}
In this paper, we proposed two boosted decision tree-based algorithms for learning STL specifications as data classifiers. In the first one, referred as offline learning, the whole batch of signals are considered for building decision trees. In the second one, referred as incremental learning, the batch of partial signals that completes incrementally over time are investigated for building the classifiers and making predictions. Our offline method shows better performance in the sense of runtime and misclassification rate compared to the literature works, and the incremental method achieves promising results that are useful for online prediction and anomaly detection on the fly.

\bibliographystyle{IEEEtran}
\bibliography{references}

\end{document}